\documentclass[]{bytedance_seed}

\usepackage[toc,page,header]{appendix}

\usepackage[utf8]{inputenc} 
\usepackage{minitoc}
\usepackage{amsfonts}
\usepackage{graphicx}
\usepackage{CJKutf8}
\usepackage{xspace}
\usepackage{todonotes}
\usepackage{enumitem}
\usepackage{fnpos} 
\usepackage{pifont}
\usepackage{booktabs}
\usepackage{array}
\usepackage{multirow}
\usepackage{xcolor}

\definecolor{Blue}{rgb}{0.255, 0.412, 0.882}
\definecolor{Green}{rgb}{0, 0.5, 0}
\colorlet{MyBlue}{Blue}  
\colorlet{MyGreen}{Green!80}

\title{Seed LiveInterpret 2.0: End-to-end Simultaneous Speech-to-speech Translation with Your Voice}

\author{ByteDance Seed}

\abstract{
Simultaneous Interpretation (SI) represents one of the most daunting frontiers in the translation industry, with product-level automatic systems long plagued by intractable challenges: subpar transcription and translation quality, lack of real-time speech generation, multi-speaker confusion, and translated speech inflation, especially in long-form discourses.
In this study, we introduce~\method, an end-to-end SI model that delivers high-fidelity, ultra-low-latency speech-to-speech generation with voice cloning capabilities. As a fully operational product-level solution,~\method tackles these challenges head-on through our novel duplex speech-to-speech understanding-generating framework. Experimental results demonstrate that through large-scale pretraining and reinforcement learning, the model achieves a significantly better balance between translation accuracy and latency, validated by human interpreters to exceed 70\% correctness in complex scenarios. Notably,~\method outperforms commercial SI solutions by significant margins in translation quality, while slashing the average latency of cloned speech from nearly 10 seconds to a near-real-time 3 seconds, which is around a near 70\% reduction that drastically enhances practical usability.
}

\date{\today}

\checkdata[Offical Page]{\url{https://seed.bytedance.com/seed_liveinterpret}}

\newcommand{\method}{\texttt{Seed LiveInterpret 2.0}\xspace}
\newcommand{\benchmark}{\texttt{RealSI}\xspace}

\newcommand{\you}{\texttt{Commercial-Y}\xspace}
\newcommand{\tong}{\texttt{Commercial-T}\xspace}
\newcommand{\bai}{\texttt{Commercial-B}\xspace}
\newcommand{\xun}{\texttt{Commercial-I}\xspace}

\newcommand{\seamless}{\texttt{SeamlessStreaming}\xspace}

\definecolor{seedblue}{HTML}{2E5AA8}

\usepackage[utf8]{inputenc} 
\usepackage{titlesec}
\usepackage{hyperref}   
\newcounter{subsubsubsection}[subsubsection]
\renewcommand\thesubsubsubsection{\thesubsubsection.\arabic{subsubsubsection}}

\titleclass{\subsubsubsection}{straight}[\subsubsection]

\titleformat{\subsubsubsection}[hang]
  {\normalfont\normalsize\sffamily\bfseries\color{seedblue}} 
  {\thesubsubsubsection}{1em}{}

\titlespacing*{\subsubsubsection}{0pt}{3.25ex plus 1ex minus .2ex}{1.5ex plus .2ex}

\begin{document}
\maketitle


\vspace{-.5cm}
\begin{figure}[!h]
    \centering
    \includegraphics[width=1.0\linewidth]{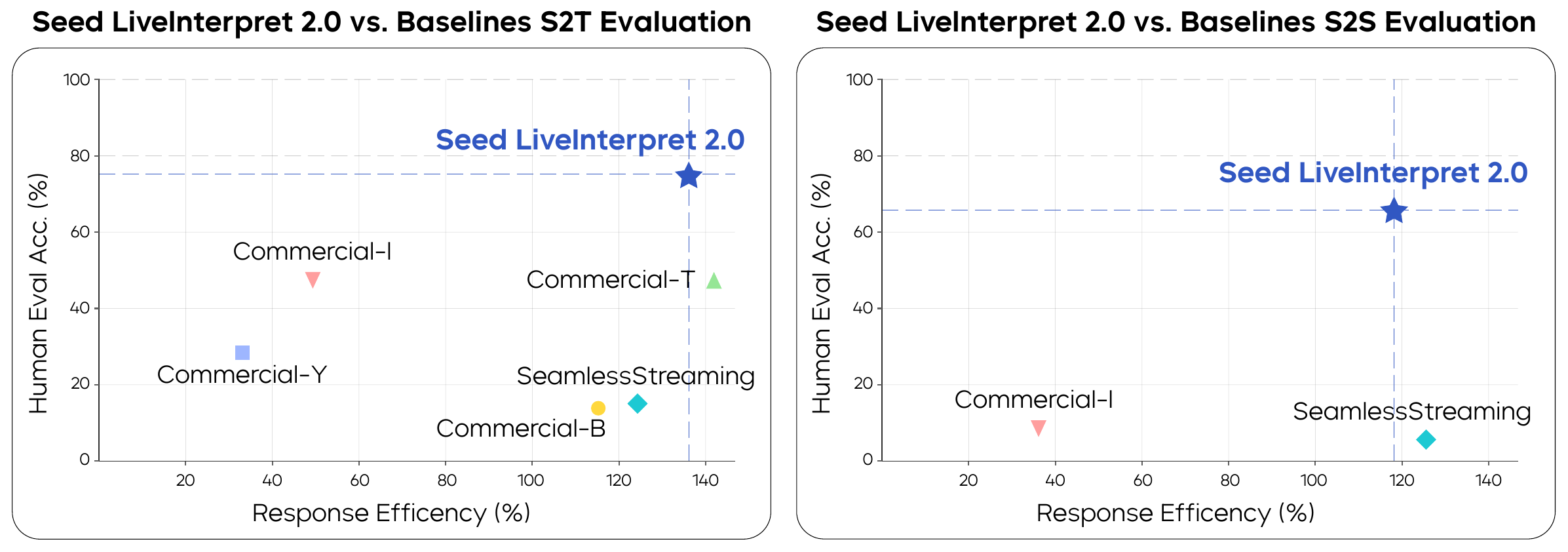}
    \vspace{-.4cm}
    \caption{Evaluation of simultaneous interpretation systems: The left and right panels compare human assessment scores of translation quality against response efficiency\protect\footnotemark for speech-to-text (S2T) and speech-to-speech (S2S) modes, where response efficiency is measured relative to human interpreter latency. 
  Human evaluation accuracy reflects how faithfully the translation output conveys the speaker’s original intent. The evaluations were conducted using the \benchmark benchmark \cite{cheng2024achievinghumanparityendtoend}. }
    \label{fig:combined}
\end{figure}
\footnotetext{Response efficiency quantifies performance relative to the latency of a human interpreter and is calculated as the quotient of~3 seconds divided by the observed latency.
}

\section{Introduction}

\begin{figure}[ht]
    \centering
    \includegraphics[width=1\linewidth]{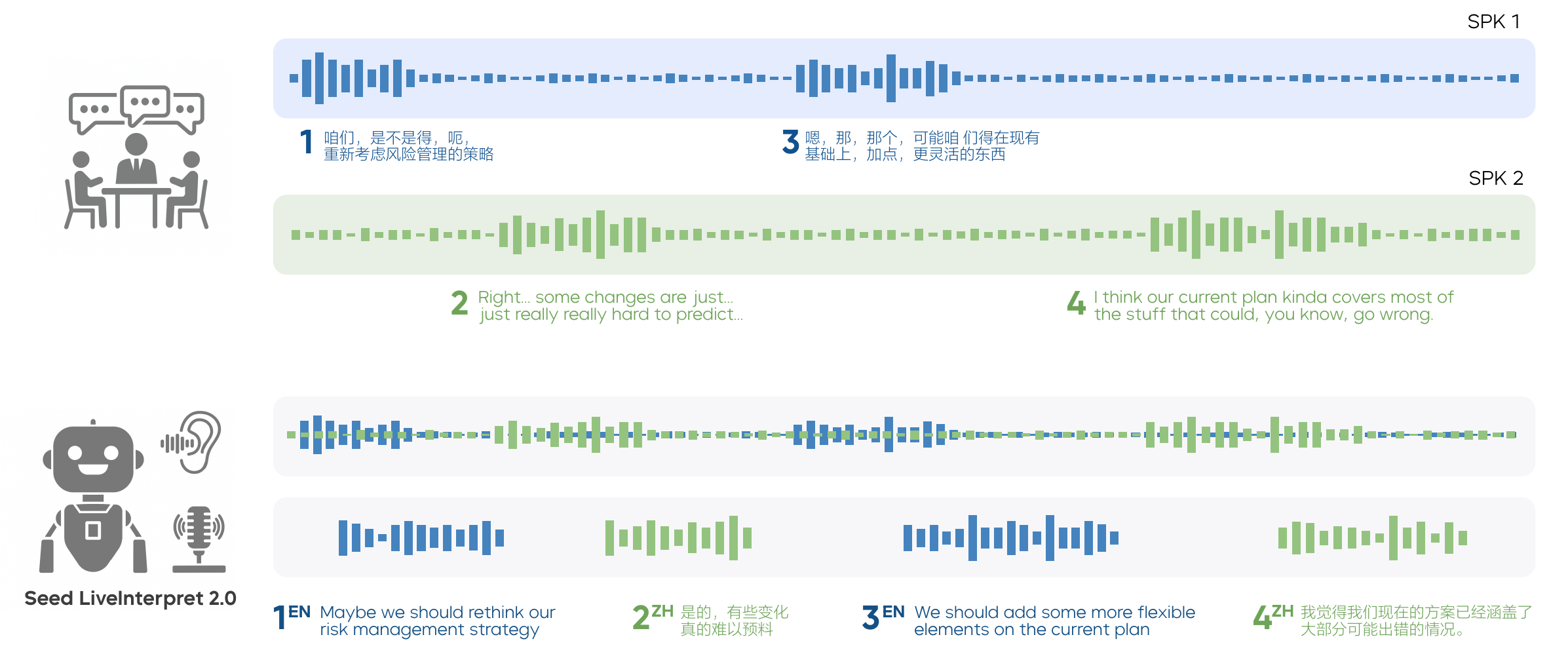}
    \caption{Illustration of \method in a multilingual live conversation scenario where two human speakers (\textcolor{MyBlue}{\textbf{SPK1}} and \textcolor{MyGreen}{\textbf{SPK2}}) communicate in Chinese and English, respectively. The top section shows the original utterances and speaker turns. Below, the \method's real-time behavior is visualized, performing simultaneous speech translation. Ear icon indicates continuous listening to each speaker. Translated outputs (in English or Chinese) appear underneath, with the horizontal gap representing translation latency. The system clones each speaker's voice and translates it into the other language using corresponding tones, represented by different colored bars. This layout highlights the system's real-time translation capabilities while maintaining speaker identity across languages.}
    \label{fig:main}
\end{figure}

Simultaneous interpretation (SI), or simultaneous speech translation\footnote{In this paper, we use the terms simultaneous interpretation and simultaneous speech translation interchangeably. For details about the \texttt{Seed LiveInterpret 1.0}, please refer to our previous technical report~\cite{cheng2024achievinghumanparityendtoend}.}, stands as one of the most challenging tasks within the translation industry~\cite{jones2014conference}. In recent years, remarkable breakthroughs have been witnessed in both machine translation and speech translation \cite{wu2016google, vaswani2017attention, nllbteam2022languageleftbehindscaling,arivazhagan2019monotonic,fukuda2023naist}. Inspired by the success of large language models (LLMs), contemporary research endeavors increasingly leverage LLMs for translation~\cite{alves2024tower,lu2024llamax,li2024eliciting,pan2024g,huang2023speech,chu2023qwen,tang2023salmonn}, aiming for superior translation performance.

However, most existing LLM-empowered speech translation systems~\cite{gaido2024speech} are limited to consecutive translation, where the model initiates translation only after the user concludes speaking, a common scenario in dialogue systems. In contrast, in contexts such as international conferences, where latency is of paramount importance to both speakers and listeners, simultaneous interpretation becomes indispensable. Current SI systems~\cite{barrault2023seamless, zhang2024streamspeech, fukuda2023naist, zeng2021realtrans, ren2020simulspeech,ouyang2025cmu,cheng2024achievinghumanparityendtoend} face significant limitations. Some rely on cascaded architectures, which are prone to error propagation and high text and speech generation latency, while others merely support end-to-end speech-to-text translation, severely restricting their practical applicability. As highlighted in~\cite{papi2025real}, the performance of existing SI systems is often overestimated due to the less stringent evaluation standards compared to offline speech translation systems. Although recent end-to-end speech-to-text models \cite{cheng2024achievinghumanparityendtoend} have improved translation quality, they still fall short in supporting low-latency, high-quality speech-to-speech translation for truly seamless interpretation. 

To tackle these challenges, we introduce \method, an end-to-end speech-to-speech simultaneous translation model that seamlessly integrates simultaneous speech-to-speech translation and voice cloning within a unified framework. As illustrated in Figure \ref{fig:main}, \method enables multilingual live conversations with natural, real-time speech translation.
The initial language model is pretrained following the methodology of the Seed LLM family~\cite{seed2025seed1,bai2024seed,anastassiou2024seed}.
We then extend this model by integrating a pre-trained audio encoder, transforming it into a multi-modal LLM capable of processing streaming audio as input. This multi-modal LLM is subsequently trained through large-scale multi-task continual learning to autoregressively generate outputs comprising text tokens (optional) and audio tokens for real-time speech synthesis~\cite{anastassiou2024seed}.
To further enhance its performance, we fine-tune the model on high-quality human-labeled data, improving its instruction following, multi-speaker discrimination, translation policy, and other critical capabilities necessary for effective simultaneous interpretation.

Recognizing the challenges of optimizing simultaneous translation under strict latency constraints, we propose a novel reinforcement learning framework for simultaneous translation that strategically balances fine-grained stepwise feedback with holistic sequence-level feedback by explicitly addressing two complementary objectives: intra-segment consistency and inter-segment coherence.
Specifically, the framework combines multi-dimensional single-turn rewards, which provide immediate feedback on translation fidelity and timing at each step to ensure intra-segment consistency, with unified multi-turn rewards that assess the overall quality and coherence of the entire output sequence, ensuring inter-segment consistency.
To address the optimization challenges arising from these complementary objectives, we adopt a two-stage training scheme: initially warming up the model by optimizing only the single-turn rewards to internalize human priors and stabilize learning, followed by further training with the multi-turn reward that jointly considers process and outcome metrics. This integrated approach enables more effective and robust reinforcement learning for simultaneous translation under strict real-time requirements.

Comprehensive experiments demonstrate that \method achieves competitive translation quality at ultra-low latency in both Chinese-to-English and English-to-Chinese translation directions, striking an optimal balance between real-time responsiveness and semantic accuracy. This research pushes the boundaries of simultaneous speech-to-speech translation by presenting a robust, natural, and end-to-end solution suitable for live applications. Our key contributions encompass a unified speech-to-speech architecture, cross-language voice cloning, and translation performance approaching human-level accuracy, as shown in Figure \ref{fig:combined}.

\section{Training}
\subsection{Continual Training and Supervised Fine-tuning}
To achieve effective modality alignment between text and speech and enhance cross-lingual capabilities, we adopt a comprehensive multitask multimodal continual training (CT) strategy. Our CT dataset encompasses nearly 100 billion tokens from diverse multimodal tasks, including audio-to-text transcription, text-to-audio synthesis, and text-only processing tasks.
Furthermore, to maximize training efficiency and ensure data quality, we employ rigorous filtering procedures based on speech quality metrics.

Following continual training, we conduct supervised fine-tuning on high-quality, human-annotated data to activate crucial capabilities required for simultaneous speech interpretation. This process enables the model to develop a data-driven read-write policy \cite{cheng2024achievinghumanparityendtoend}, multi-speaker discrimination, speech translation, and voice cloning abilities. The supervised fine-tuning significantly enhances the model's instruction-following capabilities and overall performance across essential interpretation tasks. This fine-tuned model serves as a robust foundation for subsequent reinforcement learning, enabling more targeted and effective improvements.

\subsection{Reinforcement Learning}
\subsubsection{Problem Formulation}

Modern simultaneous translation systems employ duplex processing where input streams are segmented into sequential audio chunks. Formally, we represent an input-output sequence as:
\begin{equation*}
x_{1:T} := (\text{audio}_1,y_1), (\text{audio}_2,y_2), \dots, (\text{audio}_{T}, y_{T})
\end{equation*}
where each audio chunk ($\text{audio}_t$) corresponds to incremental translation $y_t$. 
We denote $(\text{audio}_{t}, y_{t})$ as the $t$-chunk in a sequence, and $\text{audio} := (\text{audio}_{1},\text{audio}_{2},\dots \text{audio}_{T})$ as the aggregated audio from $1$ to $T$. In every $t$-chunk, we have $y_{t} := (y_t^{1},y_t^{2},\dots, y_t^{n},\dots y_t^{N})$, where $N$ is the length of the output.~\footnote{For simplicity, we denote the length of the output of each $t$-chunk as $N$. }
The model utilizes both the current audio chunk ($\text{audio}_t$) and preceding context 
$x_{<t}$ to generate translation $y_t$ through policy:
\begin{equation*}
y_{t} \sim \pi_{\theta}(\cdot|\text{audio}_{t},x_{<t}),
\end{equation*}
where $\pi_{\theta}$
is a policy with parameters $\theta$
that determines the translation strategy. The complete trajectory probability is defined as:
\begin{equation*}
\pi_{\theta}(y_{1:T}|\text{audio}) := \prod^{T}_{t=1} \pi_{\theta}(y_t|\text{audio}_{t},x_{<t}) = \prod^{T}_{t=1}\prod^N_{n=1} \pi_{\theta}(y^{n}_t|y^{<n}_{t}, \text{audio}_{t}, x_{<t}).
\end{equation*}

We denote $r^n_t$ as the reward for $n$-th token in $t$-chunk. The objective of RL is to maximize the accumulated reward along every trajectory, i.e., 
\begin{equation}
\label{sec:obj_rl}
    \mathcal{J}(\theta) = \max_\theta \, \mathbb{E}_{\substack{\text{audio} \sim \mathcal{D} \\ y \sim \pi_\theta(\cdot\,|\,\text{audio})}} 
    \left[ \sum_{t=1}^T \sum_{n=1}^{N} \gamma^{N\times t+n} r_t^n \right],
\end{equation}
where $\mathcal{D}$ is the training dataset. The following sections elaborate on how $r_t^n$ is designed.

\subsubsection{Reward Design: Balancing Single-turn and Multi-turn Feedback}

In reinforcement learning, reward mechanisms can be broadly categorized based on the temporal scope of the feedback they provide~\cite{shani2024multi}:
(1) \textit{single-turn rewards}, which provide immediate feedback assessing intermediate reasoning or generation steps at each individual decision point, and
(2) \textit{multi-turn rewards}, which evaluate the quality of the entire output sequence, reflecting long-term, cumulative outcomes across multiple decision steps.
Traditionally, reinforcement learning for language tasks often relies on reward models trained on human preference data. However, recent work has demonstrated that effective rewards can also be constructed directly from string matching, semantic similarity with labeled data~\cite{luong2024reft, openai_reinforcement}, or verifiable, rule-based criteria~\cite{lambert2024t}. These alternative approaches open new avenues for designing reward functions that do not depend solely on costly human annotations.

Simultaneous translation systems, in particular, pose distinct challenges that call for a nuanced reward design. They require optimizing two complementary objectives: (1) \textit{intra-segment consistency}, which demands that partial, real-time outputs maintain semantic and temporal integrity at each incremental step — a goal naturally suited to single-turn reward design, and (2) \textit{inter-segment coherence}, which ensures semantic and temporal consistency across the entire translated sequence — a goal addressed through multi-turn reward design that evaluates cumulative sequence-level quality. 

Motivated by these considerations, we propose a novel framework combining multi-dimensional single-turn rewards that provide fine-grained, stepwise feedback with unified multi-turn rewards that enforce global coherence and latency constraints throughout the full translation trajectory. This dual reward strategy enables more effective and balanced optimization for simultaneous translation.

\paragraph{{\textbf{Single-turn Reward}}}
\label{sec:processreward}
Motivated by recent successes in reinforcement learning with verifiable rewards~(RLVR)~\cite{luong2024reft, openai_reinforcement, guo2025deepseek, lambert2024t}, we introduce a multi-dimensional single-turn reward that evaluates both the fidelity and quality of intermediate translation steps.
Unlike traditional reward schemes focused solely on final outputs, our approach leverages granular feedback at each incremental step, which we empirically find to correlate strongly with human evaluation metrics.

Formally, given an audio sequence $\{\text{audio}_t\}_1^T$ and the corresponding ground-truth $\{y^{*}_t\}_1^T$, we define intra-segment rewards along five derived dimensions:

\begin{itemize}[leftmargin=10pt]
\item \text{Detection Accuracy Reward} ($r^{\text{l}}$): Encourage listening to avoid premature translation by penalizing outputs generated before the completion of semantic units: 
\begin{equation*}
r^{\text{l}}_{t} := \mathbb{I}(|y_t| = 0) \cdot \mathbb{I}(|y_t^*| = 0),
\end{equation*}
where $\mathbb{I}(\cdot)$ is an indicator function, and $|y_t|$ means the number of token in $y_t$.

\item \text{Translation Initiative Reward} ($r^{\text{s}}$): Encourage speech translation by rewarding the generation of confirmed semantic units as soon as they become available:
$
r^{\text{s}}_{t} := \mathbb{I}(|y_t| > 0) \cdot \mathbb{I}(|y_t^*| > 0).
$
\item \text{Translation Quality Reward} ($r^{\text{q}}$): Rewards translation quality by measuring the closeness of $y_t$ to reference $y_t$:
$r^{\text{q}}_{t} := \operatorname{Trans}(y_t, y_t^*)
$, 
where $\operatorname{Trans(\cdot, \cdot)}$ quantifies translation quality.
\item \text{Time Compliance Reward} ($r^{\text{c}}$): Encourages adherence to reference timing by rewarding generated speech durations that match the reference durations:
\begin{equation*}
r^{\operatorname{c}}_{t} := \operatorname{clip}(1 - \frac{1}{c} \max(0, \frac{\operatorname{Time}_{y_{t}} }{\operatorname{Time}_{y^*_{t}}} -1 ), -1, 1),
\end{equation*}
where $\operatorname{Time}_{y_t}$ indicates the audio duration of the translated speech, and $c$ is a constant.

\item \text{Format Consistency Reward} ($r^{\text{f}}$): Encourages the structural correctness by rewarding outputs that match a predefined pattern $\mathcal{P}$ via regular expression matching:
$\mathcal{P}$:
$r^{\text{f}}_{t} := \operatorname{RegexMatch}(y_t, \mathcal{P})$,
where $\operatorname{RegexMatch}(y_t, \mathcal{P})$  returns 1 if $y_t$ contains a substring matching the pattern $\mathcal{P}$ else 0.
\end{itemize}

Therefore, the derived multi-dimensional single-turn reward for a given audio sequence $\{\text{audio}_t\}_1^T$ is $\{r_t\}_1^T$, and the reward $r_t$ is defined as:
\begin{equation}
    r_t :=
    \begin{cases}
        w^{\text{l}}r_t^{\text{l}}, & \text{if } |y_t^*| = 0, \\
        \displaystyle\sum_{k \in \{\text{s}, \text{q}, \text{c}, \text{f}\}} w^{k} r_t^{k}, & \text{otherwise}.
    \end{cases}
\end{equation}
where $w$ are weights balancing the relative importance of each reward component.

\paragraph{\textbf{Multi-turn Reward}}
While our single-turn reward provides detailed, stepwise feedback that balances latency and translation quality at each incremental step, it does not fully capture the long-term dependencies and cumulative effects inherent in simultaneous translation. In particular, when the generated target audio increasingly lags behind the source, it causes disruptive delays that degrade user experience. To address these global sequence-level dynamics, we design a complementary multi-turn reward that evaluates the entire output sequence holistically.

This multi-turn reward enforces two critical objectives:

\begin{itemize}[leftmargin=10pt]
    \item Lagging Reward ($r^{\text{L}}$): 
    Encourages timely translation by penalizing long waiting times, and is defined as:
    $
    r^{\text{L}} := -\max\left(l, \frac{1}{K} \sum_{k=1}^K d_k \right),
    $
    where \( l \) is a reference threshold representing the maximum acceptable wait, \( K \) is the number of translation chunks, and \( d_k \) denotes the number of waited chunks before the \( k \)-th translation chunk.
    \item Sequence-level Translation Quality Reward ($r^{\text{Q}}$): Rewards the translation quality of the generated translated sequence:
    $r^{\text{Q}} := \operatorname{Align}(y, audio)$, where $\operatorname{Align}(\cdot)$ quantifies the sequence-level translation quality.
\end{itemize}

The multi-turn reward of an audio sequence is defined as:
\begin{equation*}
    r^{\text{S}} := w^\text{L}  r^{\text{L}} + w^\text{Q}  r^{\text{Q}}.
\end{equation*}
To ensure stability and comparability among reward components, each reward is normalized by subtracting its mean and dividing by its standard deviation, computed over training batches. The final reward at each time step is the sum of the normalized rewards, effectively blending local, stepwise feedback with global, sequence-level guidance.
By integrating these global constraints with fine-grained process rewards, our multi-turn reward function provides a balanced optimization signal that guides the model toward producing translations that are both timely and semantically accurate, ensuring end-to-end coherence across the full output. We also incorporate a KL divergence penalty term, $\text{KL}(\pi_\theta\|\pi_{\text{ref}})$, to regularize the learned policy toward the reference policy, promoting stable and reliable learning behavior.

\subsubsection{Stabilizing RL Training}
\label{sec:Stabilizing}
We optimize our defined objective (Eq. \ref{sec:obj_rl})
through Proximal Policy Optimization (PPO) \cite{Schulman2017ProximalPO}, which enables stable and efficient policy updates via a clipped objective function. The training objective is formulated as follows:
\begin{small}
\begin{equation*}
\mathcal{J}_{\text{PPO}}(\theta) = 
\mathbb{E}_{\substack{\text{audio} \sim \mathcal{D} \\ y \sim \pi_{\theta_{\text{old}}}}}
\left[ \sum_{n=1,t=1}^{N,T} \min \left( 
    \frac{\pi_{\theta}(y^n_{t}|\text{audio}_t,x_{<t})}{\pi_{\theta_{\text{old}}}(y^n_{t}|\text{audio}_t,x_{<t})} A^n_t,\ 
\operatorname{clip}\left(\frac{\pi_{\theta}(y^n_{t}|\text{audio}_t,x_{<t})}{\pi_{\theta_{\text{old}}}(y^n_{t}|\text{audio}_t,x_{<t})}, 1-\varepsilon, 1+\varepsilon\right) A^n_t 
\right) \right].
\end{equation*}
\end{small}
Here, $\text{audio} = \{\text{audio}_t\}_1^T$ denotes the audio sequence of the input and $y = \{y_t\}_1^T$ represents the translated response sampled from the old policy $\pi_{\theta_{old}}$.
The advantage estimate $A^n_t$ is computed using Generalized Advantage Estimation (GAE)~\cite{schulman2015high}.

Vanilla PPO underperforms in our setting because verified reward signals are prone to exploitation. For instance, when the lagging reward dominates, the model tends to produce trivial translations prioritizing latency over quality. Moreover, some rewards, such as the lagging reward, are easier to optimize than translation quality rewards, leading to an imbalance. Due to the tight coupling and diversity of these rewards, tuning their individual weights is challenging and often ineffective.
To address these issues and stabilize training, we employ two main strategies: an adaptive KL penalty \citep{Schulman2017ProximalPO,ziegler2019fine} and a two-stage reinforcement learning training scheme.




\paragraph{\textbf{Adaptive KL}} KL regularization is crucial as it constrains the policy to remain close to the reference model, thereby reducing reward hacking and preventing extreme outputs. However, controlling KL divergence is more difficult in sequences combining audio and text tokens because of their greater length, which naturally results in higher cumulative KL divergence. Consequently, the KL penalty coefficient $\beta$ must be set higher than in conventional RLHF settings. Following \citep{ziegler2019fine}, we adopt a proportional controller in log-space to adaptively adjust $\beta$, ensuring the KL divergence remains close to a predefined target.
\begin{equation*}
    \begin{aligned}
\beta_{s+1} :=\beta_s\left(1+K_\beta e_s\right), \quad e_s :=\operatorname{clip}\left(\frac{\text{KL}(\pi_\theta\|\pi_{\text{ref}})}{\mathrm{KL}_{\text {target }}}-1,-0.2,0.2\right) ,
\end{aligned}
\end{equation*}
where $s$ denotes the current training step, and $K_\beta$ is a hyper-parameter controlling the adjustment of $\beta$, and 
$\mathrm{KL}_{\text {target }}$ is a pre-defined  target $\mathrm{KL}$  divergence.

\paragraph{\textbf{Two-Stage RL Training Scheme}}
Jointly optimizing single-turn and multi-turn rewards presents a challenge: single-turn stepwise rewards are generally easier to optimize, which can cause the model to overemphasize them while neglecting sequence-level rewards that are crucial for improving overall speech translation quality.
However, these two reward types are complementary—single-turn stepwise rewards embed human priors that effectively guide early exploration, whereas multi-turn rewards drive performance refinement and global coherence.
To harness this synergy, we adopt a two-stage training scheme. In the first stage, the model is warmed up by optimizing only the multi-dimensional single-turn rewards, allowing it to internalize human priors and achieve stable learning dynamics. In the second stage, the model is further trained using the multi-turn reward that combines both process and outcome components, enabling it to refine and balance latency and translation quality effectively.
This staged approach fosters stable learning and efficient exploration, ultimately yielding a more robust and reliable reinforcement learning framework for simultaneous translation.






\begin{CJK}{UTF8}{gbsn}
\section{Experiments}
\label{sec:experiment}
\subsection{Datasets}
Our primary experiments are conducted on the recently introduced RealSI dataset \cite{cheng2024achievinghumanparityendtoend}, which encompasses both Chinese-to-English (zh-en) and English-to-Chinese (en-zh) translation directions. The dataset is sourced from a wide range of domains—including technology, healthcare, education, finance, law, environment, entertainment, science, sports, and art—and features speakers who predominantly speak naturally and casually, without extensive preparation. Each sample consists of approximately five minutes of continuous speech, providing a realistic and challenging benchmark for simultaneous interpretation systems.
Notably, the RealSI dataset reflects real-world scenarios more closely than many existing benchmarks, capturing the spontaneous and diverse nature of everyday speech across various fields.

In addition, we evaluate and compare our approach using sentence-level simultaneous translation datasets. Given the limited availability of high-quality public test data tailored for simultaneous interpretation scenarios in industry, we combine public datasets with proprietary internal datasets for evaluation.

\subsection{Evaluation Metrics}
\label{sec:metrics}
For text translation quality assessment, we primarily rely on the idea of the human evaluation metric, Valid Information Proportion (VIP) \cite{cheng2024achievinghumanparityendtoend}, which measures how accurately the translation output conveys the speaker’s original intent for each semantic fragment, closely aligning with human interpreter judgments. Additionally, we employ automated metrics such as BLEURT \cite{sellam2020bleurt} and COMET \cite{rei2020comet} as supplementary references. Nonetheless, consistent with findings in \cite{wein2024barriers, machavcek2022mt, gaido2024speech}, these automated metrics may not fully reflect the model’s true capabilities.

For speech-to-speech assessment, we propose the Speech Valid Information Proportion (SVIP)  as a comprehensive human evaluation metric.
Building upon the established Valid Information Proportion (VIP) framework \cite{cheng2024achievinghumanparityendtoend}, SVIP measures the proportion of valid speech semantic fragments within a complete speech session. A speech semantic fragment is considered valid when it effectively conveys the core information from the source speech, accurately represents the speaker's original intent, maintains delivery latency within acceptable thresholds for effective communication, sustains an appropriate pace for listener comprehension, and achieves acoustic quality that meets standards for clarity and intelligibility. SVIP provides a holistic assessment that captures not only semantic accuracy but also the pragmatic elements essential for successful spoken communication across languages. Detailed definition of SVIP can be found in Appendix~\ref{app:svip}.

For latency evaluation, we adopt the First Letter Appearance Lagging (FLAL) metric \cite{cheng2024achievinghumanparityendtoend} to measure the time until the system outputs the first determined translation at the paragraph level. At the sentence level, we use the widely adopted Average Lagging (AL) \cite{ma2018stacl} and Length Adaptive Average Lagging (LAAL) \cite{papi2022over} metrics to compare latency across different methods.

\subsection{Results}
\paragraph{\textbf{Baselines}}
We compare our \method with the open-source model SeamlessStreaming \cite{barrault2023seamless}. Due to the limited availability of baseline models, we also evaluate against several commercial systems, denoted as \texttt{Commercial-B}, \texttt{Commercial-Y}, \texttt{Commercial-T}, and \texttt{Commercial-I}. It is important to note that some baseline models employ a rewriting strategy to refine their output translations, a practice generally not used by human interpreters. In contrast, our method generates translations only when sufficient information is available, aligning more closely with the approach of human interpreters.

\begin{table*}[t]
    \centering
    \resizebox{.85\textwidth}{!}{
    \begin{tabular}{lccccccc}
        \toprule
        \multirow{2}{*}{\textbf{Model}} &   \multicolumn{3}{c}{\textbf{Speech-to-Text (zh-en)}} &   \multicolumn{4}{c}{\textbf{Speech-to-Speech (zh-en)}} \\
        \cmidrule(lr){2-4} \cmidrule(lr){5-8}
        & VIP$\uparrow$ & AL$\downarrow$ & FLAL$\downarrow$ & SVIP$\uparrow$ & AL$\downarrow$ & FLAL$\downarrow$   & Voice Clone\\
        \midrule
         {\bai} & 11.8 & 8.40 & \textcolor{white}{0}3.27  & - & - & -  & \ding{55}  \\
         {\you} & 33.2 & 3.62 & \textcolor{white}{0}5.90 & - & - & -  & \ding{55}    \\
         {\tong} & 50.1 &\bfseries  2.41 & \bfseries \textcolor{white}{0}2.35  & - & - & - & \ding{55}   \\
         {\xun} & 53.2  & 4.48 & \textcolor{white}{0}6.62   & \textcolor{white}{0}3.0 & 48.21 & 8.12 & \ding{55} \\
          {\seamless} & 22.0 & - & \textcolor{white}{0}2.65  & 15.3 & - & \bfseries 2.38   & \ding{55}\\
        \cmidrule{1-8}
         {\texttt{\textbf{Ours}}} & \bfseries 79.5 & 2.58 & \textcolor{white}{0}2.37 & \bfseries 67.8 & \bfseries  5.18 & 2.71 & \checkmark  \\
        \midrule
        \midrule
        \multirow{2}{*}{\textbf{Model}} &   \multicolumn{3}{c}{\textbf{Speech-to-Text (en-zh)}} &   \multicolumn{4}{c}{\textbf{Speech-to-Speech (en-zh)}} \\
        \cmidrule(lr){2-4} \cmidrule(lr){5-8}
        & VIP$\uparrow$ & AL$\downarrow$ & FLAL$\downarrow$ & SVIP$\uparrow$ & AL$\downarrow$ & FLAL$\downarrow$   & Voice Clone\\
        \midrule
         {\bai} & 15.5  & 4.71 & \textcolor{white}{0}1.88 & - & - &- & \ding{55}  \\
         {\you} & 24.6 & 4.96 & 12.42 & - & - & -  & \ding{55}    \\
         {\tong} & 42.0 & 2.75 & \bfseries  \textcolor{white}{0}1.90 & - & - & - & \ding{55}   \\
         {\xun} & 41.3 & 4.98 & \textcolor{white}{0}5.73   & \textcolor{white}{0}5.6 & 33.92 &8.60 &  \ding{55} \\
          {\seamless} & \textcolor{white}{0}6.0 & - & \textcolor{white}{0}2.24 & \textcolor{white}{0}2.7 & -  & 2.39 & \ding{55}\\
        \cmidrule{1-8}
         {\texttt{\textbf{Ours}}} & \bfseries 70.1  & \bfseries 2.71 & \textcolor{white}{0}2.05 & \bfseries 64.7 & \bfseries 4.75 & \bfseries 2.34  & \checkmark  \\
    \bottomrule
        
    \end{tabular}
    }
    \caption{Performance comparison on longform simultaneous translation benchmark \benchmark  across speech-to-text and speech-to-speech tasks. VIP and SVIP represent human evaluation scores for translation quality, while AL and FLAL measure translation latency at the segment level. Higher scores indicate better performance for VIP and SVIP, while lower scores are better for latency metrics. Missing entries indicate systems that do not support the corresponding functionality.}
    \label{tab:main}
\end{table*}

\paragraph{\textbf{Results on Longform Benchmark}}
Our evaluation on the longform benchmark clearly demonstrates the strengths of \method across both speech-to-text and speech-to-speech tasks, and Table~\ref{tab:main} shows the results.
For speech-to-text translation from Chinese to English (zh-en), our method achieves a human evaluation VIP score of 79.5, substantially outperforming all baselines. Alongside its superior translation quality, our approach delivers competitive latency metrics, with an AL of 2.58 and FLAL of 2.37.
Similarly, for English to Chinese (en-zh) speech-to-text translation, our model attains a VIP score of 70.1, again significantly higher than baselines such as \tong and \xun. It also achieves the best latency results, with an AL of 2.71 and FLAL of 2.05, indicating a well-balanced trade-off between translation quality and delay.

In speech-to-speech translation, our method achieves the low latency measurements, outperforming the baseline systems by a substantial margin.  This highlights the model’s ability to maintain high-quality output while minimizing translation delay.
Notably, many commercial systems either do not support speech-to-speech translation or show significantly degraded performance in longform scenarios, highlighting the practical value of our approach for real-world applications.
In speech-to-speech translation, our model demonstrates strong performance in both directions. For zh-en, it achieves the highest SVIP score of 67.8 and the lowest AL of 5.18 among systems supporting this task, with an FLAL of 2.71. For en-zh, our approach reaches an SVIP of 64.7 and improves latency significantly, with an AL of 4.75 and FLAL of 2.34, outperforming all other speech-to-speech baselines. Notably, many commercial systems either do not support speech-to-speech translation or show degraded performance in longform scenarios, emphasizing the practical advantage of our approach in real-world applications.
Importantly, our method is the only system in the comparison that supports voice cloning, enabling personalized speech output in simultaneous translation, which further enhances user experience and applicability.

Overall, these results highlight that \method consistently achieves state-of-the-art translation quality with low latency across both language pairs and modalities, confirming its effectiveness and robustness for simultaneous translation in longform settings.

\paragraph{\textbf{Results on Sentence-Level Benchmark}}
We evaluate our \method model against established baselines on sentence-level zh-en and en-zh datasets, examining the performance across both speech-to-text and speech-to-speech simultaneous translation tasks, and Table \ref{tab:translation_quality_automatic_metrics_other} shows the results.
For speech-to-text translation, our approach consistently achieves the highest translation quality across both datasets, outperforming commercial systems. Specifically, our model attains BLEURT scores of 64.9 (zh-en) and 62.0 (en-zh), along with COMET scores of 84.1 and 85.3, respectively. These results demonstrate a clear advantage in translation accuracy.
In terms of latency, our method achieves the lowest AL on zh-en and competitive AL on en-zh, indicating faster translation without sacrificing quality. While \tong achieves slightly better FLAL on zh-en, it does so at the cost of significantly lower BLEURT and COMET scores. Similarly, other commercial systems such as \seamless and \bai exhibit trade-offs between latency and quality, whereas our model maintains a strong balance between both.

For speech-to-speech translation, our method also leads in translation quality, with BLEURT scores of 60.7 and 57.6 and COMET scores of 83.6 and 83.5, respectively. Although its latency metrics are slightly higher than the lowest values reported by some baselines, our approach consistently achieves a favorable trade-off by delivering superior translation quality alongside competitive latency.
Overall, these results highlight the ability of \method to effectively balance translation quality and latency across both speech-to-text and speech-to-speech tasks, outperforming existing commercial systems on sentence-level benchmarks.

\begin{table*}[t]
    \centering
    \resizebox{1.0\textwidth}{!}{
    \begin{tabular}{lcccccccccc}
        \toprule
        \multirow{2}{*}{\textbf{Model}} & \multicolumn{5}{c}{\textbf{Speech to Text (zh-en)}} & \multicolumn{5}{c}{\textbf{Speech to Speech (zh-en)}} \\
        \cmidrule(lr){2-6} \cmidrule(lr){7-11} &
        BLEURT$\uparrow$ & COMET$\uparrow$ & AL$\downarrow$ & LAAL$\downarrow$ & FLAL$\downarrow$ & BLEURT$\uparrow$ & COMET$\uparrow$ & AL$\downarrow$ & LAAL$\downarrow$ & FLAL$\downarrow$  \\
        \midrule
         {\you} & 61.5 & 81.3 & 2.03 & 2.11 & 1.80 & - & - & - & - & - \\
         {\tong} & 61.9 & 81.5 & 1.61 & 1.75 & \bfseries 1.74 & - & - & - & - & -  \\
         {\bai} & 47.2 & 70.3  & 2.39 & 2.66 & 2.21 & 44.8 & 71.6 & 12.00 & 12.26 & 7.89 \\
         {\xun} & 55.9 & 79.0 & 3.10 & 3.22 & 4.62 & 53.2 & 79.6 & 6.90 & 7.02 & 4.73  \\
         {\seamless} & 55.8 & 76.4 &  1.68 & 1.87 & 2.36 & 49.6 & 75.2 & \bfseries  2.96 & \bfseries  3.10 & \bfseries  2.53 \\
        \cmidrule(lr){1-11}
         {\texttt{\textbf{Ours}}} & \bfseries 64.9 & \bfseries 84.1 & \bfseries 1.37 & \bfseries 1.56 & 2.12 & \bfseries 60.7 & \bfseries  83.6 & 3.56 & 3.79 & 3.08 \\
        \midrule
        \midrule
        \multirow{2}{*}{\textbf{Model}} & \multicolumn{5}{c}{\textbf{Speech to Text (en-zh)}} & \multicolumn{5}{c}{\textbf{ Speech to Speech (en-zh)}} \\
        \cmidrule(lr){2-6} \cmidrule(lr){7-11} &
        BLEURT$\uparrow$ & COMET$\uparrow$ & AL$\downarrow$ & LAAL$\downarrow$ & FLAL$\downarrow$ & BLEURT$\uparrow$ & COMET$\uparrow$ & AL$\downarrow$ & LAAL$\downarrow$ & FLAL$\downarrow$ \\
        \midrule
         {\you} & 59.5 & 83.1 & 3.25 & 3.51 & 4.84 & - & - & - & - & -  \\
         {\tong} &  60.1 & 84.1 & \bfseries  1.46 & \bfseries  1.71 & \bfseries  1.51 & - & - & - & - & -  \\
         {\bai} & 55.2 & 81.2 & 2.62 & 2.91 & 2.07 &  49.0 & 77.7 & 13.10 & 13.57 & 8.91 \\
         {\xun} & 60.0 & 83.4 & 3.25 & 3.54 & 4.86 &  56.1 & 82.1 & 7.25 & 7.58 & 5.49 \\
         {\seamless} & 48.2 & 75.2 & 1.43 & 1.69 & 2.06 & 40.4 & 69.8 & 3.30 & 3.69 & \bfseries 2.17  \\
        \cmidrule(lr){1-11}
        {\texttt{\textbf{Ours}}} & \bfseries 62.0 & \bfseries 85.3 & 2.17 & 2.18 & 2.28 & \bfseries 57.6 & \bfseries 83.5 & \bfseries 2.81 & \bfseries 3.12 & 2.38  \\
        \bottomrule
    \end{tabular}} 
    \caption{Comparisons of our method with baseline approaches on speech-to-text and speech-to-speech simultaneous translation tasks with respect to translation quality and latency on the sentence-level datasets.}
    \label{tab:translation_quality_automatic_metrics_other}
\end{table*}
\end{CJK}
\section{Analysis}
\begin{table*}[t]
    \centering
    \resizebox{0.6\textwidth}{!}{
    \begin{tabular}{lccccccc}
        \toprule
        \multirow{2}{*}{\textbf{Model}} &   \multicolumn{3}{c}{\textbf{Speech-to-Text (zh-en)}} &   \multicolumn{3}{c}{\textbf{Speech-to-Speech (zh-en)}} \\
        \cmidrule(lr){2-4} \cmidrule(lr){5-7}
        & VIP$\uparrow$ & AL$\downarrow$ & FLAL$\downarrow$ & SVIP$\uparrow$  & AL$\downarrow$ &  FLAL$\downarrow$   \\
        \midrule
         \texttt{Ours$_\text{(SFT)}$} & 75.1 & 2.82 & 3.90  & 66.6 & 5.80  & 4.26  \\
         \texttt{Ours} & \bfseries 79.5 & \bfseries 2.58 & \bfseries 2.37 & \bfseries 67.8 & \bfseries 5.18 & \bfseries 2.71 \\
        \bottomrule
    \end{tabular}
    }
    \caption{Performance comparison on longform simultaneous translation benchmark \benchmark.}
    \label{tab:sft_rl_vip}
\end{table*}
\subsection{Comparisons of SFT with RL}
Tables~\ref{tab:sft_rl_vip} and~\ref{tab:sft_rl_auto} compare our \method with its
SFT version across benchmarks and tasks.
On the longform simultaneous translation benchmark (\benchmark), both our model and its SFT version demonstrate strong performance across speech-to-text and speech-to-speech tasks. Notably, our method achieves a higher human evaluation VIP score for speech-to-text translation (79.5 vs. 75.1), while significantly reducing latency. Specifically, FLAL decreases from 3.90 to 2.37, and AL improves from 2.82 to 2.58, indicating faster and more responsive translations with minimal quality trade-off.
In speech-to-speech translation, our method maintains competitive SVIP scores (67.8 vs. 66.6) while substantially lowering latency metrics. The FLAL metric drops from 4.26 to 2.71, and AL reduces from 5.80 to 5.18, underscoring our model’s ability to deliver timely and high-quality speech output.

On sentence-level datasets, our method consistently outperforms the SFT model in translation quality, achieving BLEURT improvements of 0.4 to 1.0 points alongside modest gains in COMET scores. More importantly, our approach delivers significant latency reductions across key metrics, including AL, LAAL, and FLAL. For instance, in speech-to-text translation (zh-en), AL decreases from 1.69 to 1.37 and FLAL from 3.03 to 2.12; similarly, for speech-to-text (en-zh), AL is reduced from 2.99 to 2.17 and FLAL from 3.29 to 2.28. In speech-to-speech tasks, our model maintains comparable or slightly improved translation quality while substantially lowering latency, exemplified by a reduction in AL from 3.99 to 2.81.

Overall, our \method method consistently optimizes the trade-off between translation quality and latency across datasets and tasks. While quality improvements are moderate, the significant latency reductions highlight the effectiveness of reinforcement learning in producing faster, high-quality simultaneous translation.

\begin{table*}[t]
    \centering
    \resizebox{1.0\textwidth}{!}{
    \begin{tabular}{lcccccccccc}
        \toprule
        \multirow{2}{*}{\textbf{Model}} & \multicolumn{5}{c}{\textbf{Speech to Text (zh-en)}} & \multicolumn{5}{c}{\textbf{Speech to Speech (zh-en)}} \\
        \cmidrule(lr){2-6} \cmidrule(lr){7-11} &
        BLEURT$\uparrow$ & COMET$\uparrow$ & AL$\downarrow$ & LAAL$\downarrow$ & FLAL$\downarrow$ & BLEURT$\uparrow$ & COMET$\uparrow$ & AL$\downarrow$ & LAAL$\downarrow$ & FLAL$\downarrow$  \\
        \midrule
         \texttt{Ours$_\text{(SFT)}$} & 64.5 & 83.9 & 1.69 & 1.91 & 3.03 & 59.7 & 83.2 & 3.57 & 3.80 & 3.08  \\
         \texttt{Ours} & \bfseries 64.9 & \bfseries 84.1 & \bfseries 1.37 & \bfseries 1.56 & \bfseries2.12 & \bfseries 60.7 & \bfseries  83.6 & \bfseries 3.56 & \bfseries 3.79 & \bfseries  3.08 \\
        \midrule
        \midrule
        \multirow{2}{*}{\textbf{Model}} & \multicolumn{5}{c}{\textbf{Speech to Text (en-zh)}} & \multicolumn{5}{c}{\textbf{ Speech to Speech (en-zh)}} \\
        \cmidrule(lr){2-6} \cmidrule(lr){7-11} &
        BLEURT$\uparrow$ & COMET$\uparrow$ & AL$\downarrow$ & LAAL$\downarrow$ & FLAL$\downarrow$ & BLEURT$\uparrow$ & COMET$\uparrow$ & AL$\downarrow$ & LAAL$\downarrow$ & FLAL$\downarrow$  \\
        \midrule
         \texttt{Ours$_\text{(SFT)}$} & 61.0 & 84.7 & 2.99 & 3.01 & 3.29 & \bfseries 57.6 & \bfseries 83.5 & 3.99 & 4.32 & 3.40  \\
        \texttt{Ours} & \bfseries 62.0 & \bfseries 85.3 & \bfseries 2.17 & \bfseries 2.18 & \bfseries 2.28 & \bfseries 57.6 & \bfseries 83.5 & \bfseries 2.81 & \bfseries 3.12 & \bfseries 2.38 \\
        \bottomrule
    \end{tabular}} 
    \caption{Comparisons of \method with baseline approaches on speech-to-text and speech-to-speech simultaneous translation tasks with respect to translation quality and latency on the sentence-level datasets.}
    \label{tab:sft_rl_auto}
\end{table*}

\subsection{Balanced Reward to Prevent Hacking}
Despite applying the stabilization techniques described in Section \ref{sec:Stabilizing}, we observe that our model remains vulnerable to reward hacking, even when using verifiable rewards. Table \ref{tab:rewardhack} illustrates this phenomenon with respect to the Time Compliance Reward ($r^\text{c}$) defined in Section \ref{sec:processreward}, which is designed to encourage the generated speech durations to adhere to the reference source durations.
When training the model solely with the 
$r^\text{c}$ reward, we find that the model exploits this signal by significantly reducing the duration of the generated audio — approximately a 35\% decrease on both the en-zh and zh-en datasets. Correspondingly, the number of generated text tokens also declines by about 15\%. This reduction leads to a substantial drop in BLEURT scores, indicating degraded translation quality.
This behavior suggests that while the model successfully aligns the duration of generated speech with the source audio (thus maximizing the 
$r^\text{c}$ reward), it does so at the cost of omitting substantial semantic content. In other words, the model learns to produce shorter outputs that satisfy the temporal constraint but sacrifice translation fidelity.

Therefore, we find that it is necessary to add an adversarial quality reward ($r^\text{q}$) alongside the
$r^\text{c}$
reward to balance translation quality with temporal constraints. With this combined reward scheme, the model maintains a comparable number of text tokens while reducing the duration of the audio by only about 15\%. Importantly, BLEURT scores remain stable, indicating preserved translation quality.
These results suggest that incorporating adversarial or complementary rewards is essential to prevent reward hacking and encourage balanced optimization across multiple objectives.

\begin{table*}[h]
    \centering
    \resizebox{1\textwidth}{!}{
    \begin{tabular}{lccccccc}
        \toprule
        \multirow{2}{*}{\textbf{Model}} & \multirow{2}{*}{\textbf{Reward}} & \multicolumn{3}{c}{\textbf{Chinese $\to$ English}} & \multicolumn{3}{c}{\textbf{English $\to$ Chinese}} \\
        \cmidrule(lr){3-5} \cmidrule(lr){6-8}
        & & {Text Length} & {Audio Duration} & BLEURT & {Text Length} & {Audio Duration} & BLEURT \\
        \midrule
        \texttt{Ours$_\text{(SFT)}$} & - & 113,000 & 9,340 & 56.36 & 23,000 & 2,080 & 58.00 \\
        \texttt{Ours} & $r^{\text{c}}$ & \textcolor{white}{0}97,000 & 6,350 & 48.31 & 21,000 & 1,330 & 46.43 \\
        \texttt{Ours} & $r^{\text{c}} + r^{\text{q}}$ & 114,000 & 8,010 & 55.60 & 24,000 & 1,760 & 58.07 \\
        \bottomrule
    \end{tabular}
    }
    \caption{
    Ablation study of reward configurations across both translation directions on in-house datasets.
    Text length indicates the total number of generated tokens, while audio duration shows the total speech length (in seconds) across the dataset.
    Incorporating adversarial or complementary rewards helps prevent reward hacking and encourages balanced optimization across multiple objectives.
    }
    \label{tab:rewardhack}
\end{table*}

\subsection{Effect of Two-Stage RL Training Scheme}

We investigate the impact of single-turn and multi-turn rewards on the performance of our RL model through an ablation study, comparing three training configurations. Table \ref{tab:reward_combined_simple} presents the results. 
The first uses only single-turn rewards~($r_t$), the second relies solely on multi-turn rewards ($r^{\text{S}}$), and the third employs our proposed two-stage training scheme combining both.
Compared to the single-turn-only approach (\texttt{Ours$_\text{(single)}$}), the multi-turn variant (\texttt{Ours$_\text{(multi)}$}) significantly reduces latency—2.20 versus 2.62 for zh-en, and 3.16 versus 2.69 for English-to-Chinese. However, it achieves lower VIP scores, indicating a tendency to exploit the lagging reward ($r^{\text{L}}$) at the expense of translation quality. This trade-off is undesirable, as it favors speed over output accuracy. Conversely, the single-turn-only method exhibits relatively high latency, suggesting that focusing exclusively on process rewards limits the model’s ability to explore more efficient translation strategies.

Our \method that employs a two-stage training strategy outperforms both single-turn and multi-turn models across key metrics. 
It achieves competitive translation quality while maintaining latency close to that of the outcome-only setup. Notably, in the en-zh direction, the two-stage model attains a high VIP score with reduced lagging, demonstrating that combining process and outcome rewards effectively balances translation quality and timeliness.

\begin{table*}[h]
    \centering
    \resizebox{0.85\textwidth}{!}{
    \begin{tabular}{lcccccccc}
        \toprule
        \multirow{2}{*}{\textbf{Model}} 
        & \multicolumn{4}{c}{\textbf{Chinese $\to$ English}} 
        & \multicolumn{4}{c}{\textbf{English $\to$ Chinese}} \\
        \cmidrule(lr){2-5} \cmidrule(lr){6-9}
        & $\text{BLEURT}\uparrow$ & $\text{COMET}\uparrow$ & $\text{VIP}\uparrow$ & $\text{AL}\downarrow$ 
        &$\text{BLEURT}\uparrow$ & $\text{COMET}\uparrow$ & $\text{VIP}\uparrow$ & $\text{AL}\downarrow$\\
        \midrule
        \texttt{Ours$_\text{(single)}$} & 55.83 & 80.14 & 70\% & 2.62 & 57.08 & 85.47 & 76\% & 3.16 \\
        \texttt{Ours$_\text{(multi)}$} & 54.05 & 78.46 & 67\% & 2.20 & 58.35 & 85.70 & 72\% & 2.69 \\
        \texttt{Ours} & 55.60 & 80.09 & 71\% & 2.30 & 58.07 & 85.60 & 76\% & 2.78 \\
        \bottomrule
    \end{tabular}
    }
    \caption{Ablation study of different reward strategies across both translation directions on in-house datasets.}
    \label{tab:reward_combined_simple}
\end{table*}

\section{Related Work}
Simultaneous Interpretation aims to translate spoken language in real time, facilitating seamless multilingual communication. Traditional SI systems typically rely on cascaded architectures that sequentially perform automatic speech recognition, machine translation, and text-to-speech synthesis \cite{gu2016learning, cho2016can}. While these modular pipelines allow for targeted optimization at each stage, they suffer from error propagation and increased latency, which negatively impact overall translation quality and responsiveness. To mitigate these issues, recent research \cite{jia2019direct, zeng2021realtrans, ren2020simulspeech} has shifted towards end-to-end models that directly convert source speech into translated text or speech, thereby reducing latency and minimizing accumulated errors. However, most existing end-to-end SI frameworks \cite{cheng2024achievinghumanparityendtoend,  zhang2024streamspeech, barrault2023seamless} focus predominantly on speech-to-text translation and lack comprehensive support for speech-to-speech translation with speaker voice cloning, an essential feature for preserving speaker identity and enhancing user experience. Achieving the trifecta of low latency, high fidelity, and voice preservation remains a critical challenge in this domain.

Simultaneous translation methods generally adopt either fixed or adaptive policies for deciding when to read input and emit output. Fixed policy approaches follow predefined rules, such as reading a fixed number of source tokens before generating target tokens~\citep{ma2018stacl, elbayad2020efficient}. These methods are simple but inflexible, often resulting in suboptimal latency-quality trade-offs. Adaptive policies dynamically determine read/write actions based on contextual alignment between source and target sequences, allowing for more nuanced and responsive translation~\citep{zheng2019simpler, arivazhagan2019monotonic, liu2021cross, zhang2023hidden, barrault2023seamless, cheng2024achievinghumanparityendtoend}. However, adaptive approaches rely heavily on alignment signals that can be noisy and challenging to optimize without reinforcement learning~\citep{xu2025seqpo}. Recent work has leveraged RL to learn more effective read/write policies, overcoming limitations of rule-based and alignment-dependent methods~\citep{yu2025simulpl, xu2025seqpo}. Despite these advances, RL-based methods have been primarily applied to text-to-text translation, with speech-to-speech simultaneous translation remaining largely unexplored.

Reinforcement learning has been widely employed to improve offline machine translation quality. Techniques such as reinforcement learning from human feedback have enhanced translation fluency, adequacy, and alignment with human preferences \cite{xu2024advancing, feng2025mt, he2025r1, he2024improving}. Nonetheless, these approaches focus on text input and output and do not address the challenges of real-time or speech-to-speech translation. The most relevant RL-based simultaneous translation work \citep{yu2025simulpl, xu2025seqpo} concentrates on real-time text translation policies and does not extend to the full end-to-end speech-to-speech pipeline, which introduces additional complexities such as speech synthesis and voice cloning.

\section{Conclusion}

In this work, we have presented \method, an end-to-end speech-to-speech simultaneous translation approach that seamlessly integrates translation, voice cloning, and speech synthesis within a unified framework. By leveraging a duplex processing architecture and a multimodal large language model, our approach achieves ultra-low latency and natural cross-language voice cloning, addressing key limitations of prior cascaded and end-to-end models. We introduce a novel two-stage reinforcement learning framework that employs a unified reward design that balances fine-grained process-based feedback with global outcome-based objectives, enabling the model to optimize both translation quality and latency in real time.
Extensive experiments demonstrate that our approach achieves competitive translation accuracy while maintaining stringent latency requirements, paving the way for more robust and natural simultaneous interpretation systems applicable in live multilingual communication scenarios. Future work will explore further improvements in voice personalization, speech stability, expressiveness, as well as scaling to a broader range of languages and acoustic conditions.

\clearpage

\bibliographystyle{plainnat}
\bibliography{main}

\clearpage

\beginappendix

\section{Guidelines of SVIP}
\label{app:svip}
The SVIP assessment uses a two-step evaluation process. First, language quality is evaluated using VIP metric~\cite{cheng2024achievinghumanparityendtoend}. If the VIP score is zero, the SVIP score is automatically zero. For non-zero VIP scores, speech quality indicators (listed in Table~\ref{tab:svip}) determine the final SVIP score: it is zero if any indicator scores 1 point, and it is 1 if all indicators score 3 points or higher. When some indicators score 2 points, the final decision depends on whether the overall message remains comprehensible at the sentence level.

We define the \textbf{speech quality indicators} as in Table~\ref{tab:svip}:

\begin{table}[h!]
\centering
\small
\renewcommand{\arraystretch}{1.3}
\begin{tabular}{p{2.2cm}p{6.5cm}p{6.3cm}}
\toprule
\textbf{Indicator} & \textbf{Description} & \textbf{Scoring Criteria} \\
\midrule

\multirow{5}{2.2cm}{\textbf{Latency}} & \multirow{5}{6.5cm}{The time interval from when AI or human interpreters receive the original audio to when they output the interpreted speech, evaluating the real-time responsiveness of interpretation. Low latency ensures smooth communication flow and coherence.} & 
\textbf{5:} Very short latency (<3 seconds) \\
& & \textbf{4:} Moderate latency (3-5 seconds) \\
& & \textbf{3:} Longer latency (5-7 seconds) \\
& & \textbf{2:} Extended latency (7-9 seconds) \\
& & \textbf{1:} Excessive latency ($\geq$10 seconds) \\
\midrule

\multirow{5}{2.2cm}{\textbf{Speech Rate}} & \multirow{5}{6.5cm}{Whether the interpretation speech rate is appropriate for audience comprehension and language habits, ensuring smooth rhythm and rapid and effective information transmission while avoiding comprehension difficulties due to excessive speed or attention distraction due to excessive slowness, ensuring the audience can efficiently and smoothly obtain interpretation content.} & 
\textbf{5 :} Moderate speed, natural \\
& & \textbf{4:} Slightly fast or slow but acceptable \\
& & \textbf{3:} Obviously too fast or slow, affecting experience but not comprehension \\
& & \textbf{2:} Obviously too fast or slow, affecting content comprehension \\
& & \textbf{1:} Obviously too fast or slow, seriously affecting content comprehension \\
\midrule

\multirow{5}{2.2cm}{\textbf{Pronunciation}} & \multirow{5}{6.5cm}{The accuracy and clarity of pronunciation in simultaneous interpretation, evaluating whether the speech expression is clear and comprehensible, without obvious reading or stuttering situations, and the degree to which pronunciation quality affects audience understanding of information.} & 
\textbf{5:} Completely accurate pronunciation \\
& & \textbf{4:} Minor pronunciation issues, acceptable \\
& & \textbf{3:} Obvious pronunciation errors, affecting experience but not comprehension \\
& & \textbf{2:} Obvious pronunciation errors, affecting content comprehension \\
& & \textbf{1:} Obvious pronunciation errors, seriously affecting content comprehension \\
\midrule

\multirow{5}{2.2cm}{\textbf{Fluency}} & \multirow{5}{6.5cm}{Whether there are phenomena affecting listening experience in speech output, and whether the overall speech rhythm meets audience listening habits. Good speech fluency should show continuous and natural expression, without obvious interruptions or stuttering.} & 
\textbf{5:} High fluency, coherent expression, no obvious pauses, repetitions, or hesitations \\
& & \textbf{4:} Occasional short pauses or minor hesitations, overall good speech flow and rhythm \\
& & \textbf{3:} Obvious fluency issues, affecting rhythm but basically acceptable \\
& & \textbf{2:} Obvious stuck or stuttering, significantly reduced fluency, affecting comprehension \\
& & \textbf{1:} Serious fluency issues, basically incomprehensible \\

\bottomrule
\end{tabular}
\caption{Detailed descriptions of speech quality indicators.}
\label{tab:svip}
\end{table}

\newpage

\section{Contributions}

\textbf{Project Lead}

Shanbo Cheng (\email{chengshanbo@bytedance.com})

\textbf{Core Contributors}

Yu Bao, Zhichao Huang, Yu Lu, Ningxin Peng, Lu Xu, Runsheng Yu

\textbf{Contributors}

Rong Cao, Yujiao Du, Ting Han, Yuxiang Hu, Zeyang Li, Sitong Liu, Shengtao Ma, Shiguang Pan, Jiongchen Xiao, Nuo Xu, Meng Yang, Rong Ye, Yiming Yu, Jun Zhang, Ruofei Zhang, Wanyi Zhang, Wenhao Zhu, Liehao Zou

\textbf{Acknowledgement}

Jinping Cai, Pengli Chen, Zhuo Chen, Qianqian Dong, Meng Ge, Minglun Han, Shan He, Xiaomin Huang, Youjia Huang, Yuanyuan Huo, Fanliu Kong,  Hang Li, Mengnan Li, Xiaoyang Li, Xingxing Li, Yifu Li, Shouda Liu, Xinyu Liu, Xiaoying Jia, Yongjian Mao, Junjie Pan, Xinghua Qu, Hongbin Ren, Chen Shen, Kefei Sun, Tian Tan, Ming Tu, Bo Wang, Yuping Wang, Manlian Wu, Hanzhang Xia, Bowen Xiao, Yangfei Xu, Bing Yang, Yu Yang, Bairen Yi, Yang Zhang

\textbf{Supervision}

Lu Lu, Yuxuan Wang, Yonghui Wu

\vspace{0.5cm}
We would also like to sincerely thank all our colleagues for their invaluable support.

\vspace{0.5cm}
(Last-Name in Alphabetical Order)

\end{document}